\def\year{2019}\relax
\documentclass[letterpaper]{article} 
\usepackage{aaai19}  
\usepackage{times}  
\usepackage{helvet}  
\usepackage{courier}  
\usepackage{url}  
\usepackage{graphicx}  
\usepackage{amsmath,amssymb}
\usepackage{xcolor}
\usepackage{comment}
\usepackage{multirow}
\usepackage{tabularx}
\begin{document}
%
\title{Gradient Harmonized Single-stage Detector}
\author{Buyu Li\thanks{They contributed equally to this work}, Yu Liu\footnotemark[1] and Xiaogang Wang\\
\{byli, yuliu, xgwang\}@ee.cuhk.edu.hk\\
Multimedia Laboratory, The Chinese University of Hong Kong, Hong Kong\\
}
\maketitle
\begin{abstract}
Despite the great success of two-stage detectors, single-stage detector is still a more elegant and efficient way, yet suffers from the two well-known disharmonies during training, i.e. the huge difference in quantity between positive and negative examples as well as between easy and hard examples. In this work, we first point out that the essential effect of the two disharmonies can be summarized in term of the gradient. Further, we propose a novel gradient harmonizing mechanism (GHM) to be a hedging for the disharmonies. The philosophy behind GHM can be easily embedded into both classification loss function like cross-entropy (CE) and regression loss function like smooth-$L_1$ ($SL_1$) loss. To this end, two novel loss functions called GHM-C and GHM-R are designed to balancing the gradient flow for anchor classification and bounding box refinement, respectively. Ablation study on MS COCO demonstrates that without laborious hyper-parameter tuning, both GHM-C and GHM-R can bring substantial improvement for single-stage detector. Without any whistles and bells, the proposed model achieves 41.6 mAP on COCO \textit{test-dev} set which surpass the state-of-the-art method, Focal Loss (FL) + $SL_1$, by 0.8. The code\footnote{https://github.com/libuyu/GHM\_Detection} is released to facilitate future research. 
\end{abstract}

\section{Introduction}

One-stage approach is the most efficient and elegant framework for object detection. But for a long time, the performance of one-stage detectors has a large gap from that of two-stage detectors. The most challenging problem for the training of one-stage detector is the serious imbalance between easy and hard examples as well as that between positive and negative examples. The huge number of easy and background examples tend to overwhelm the training. But these problems are not existed for two-stage detectors, owing to the proposal-driven mechanism. To handle the former imbalance problem, example mining based methods such as OHEM \cite{ohem} are in common use, but they directly abandon most examples and the training is inefficient. For the latter imbalance, the recent work, Focal Loss \cite{focal}, has tried to address it by rectifying the cross-entropy loss function to a elaborately designed form. However, Focal Loss adopts two hyper-parameters which should be tuned with a lot of efforts. And it is a static loss which is not adaptive for the changing of data distribution, which varies along with the training process.

\begin{figure}[t]
\centering
\includegraphics[width=1\linewidth]{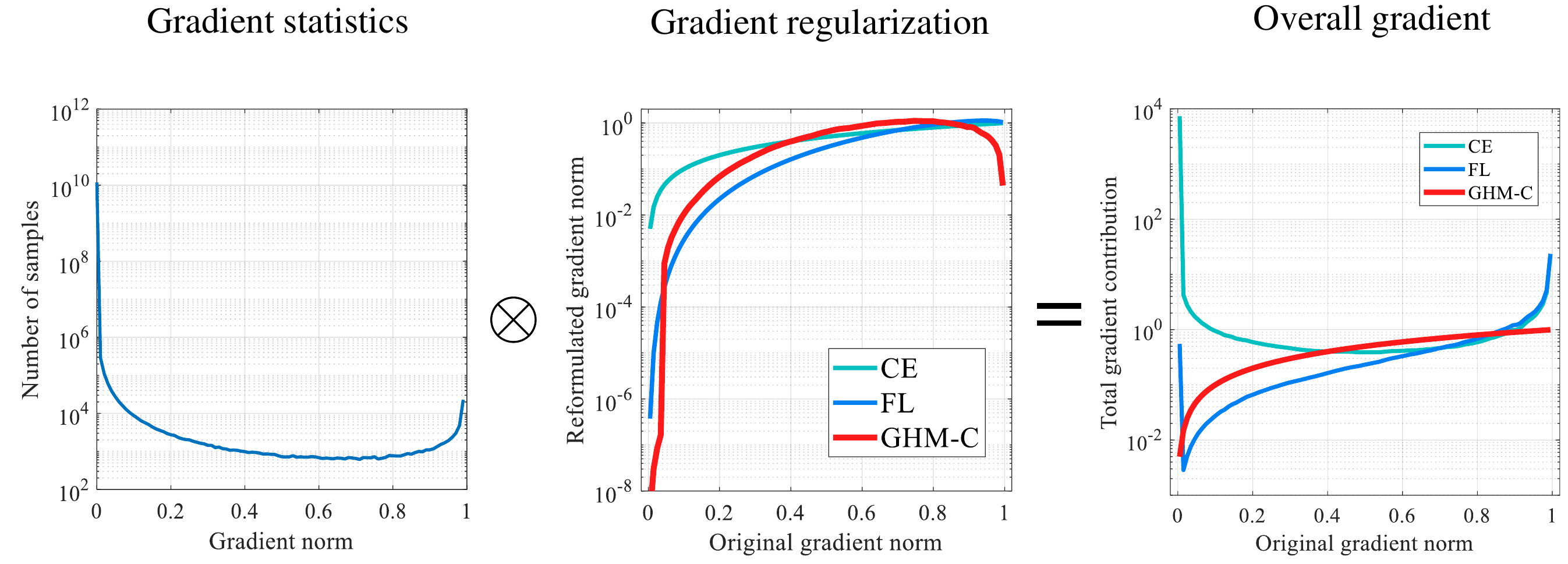}
\caption{An illustration of gradient harmonizing mechanism. The figure in the left displays the distribution of relative gradient norm in a converged model in log scale respectively. The middle figure displays the new gradient norms after the rectification of Focal Loss (FL) and GHM-C loss, compared with the original cross-entropy (CE) loss. The right figure shows the total gradient contribution of examples w.r.t gradient norm.}
\label{fig:method}
\end{figure}

In this work, we first point out that the class imbalance can be summarized to the imbalance in difficulty and the imbalance in difficulty can be summarized to the imbalance in gradient norm distribution. If a positive example is well-classified, it is an easy example and the model benefit little from it, i.e. a little magnitude of gradient will be produced by this sample. And a misclassified example should attract attention of the model no matter which class it belongs to. So if viewed globally, the large amount of negative examples tends to be easy to classify and the hard examples are usually positive. So the two kind of imbalance can be roughly summed up as attribute imbalance.

Moreover, we claim that the imbalance of examples with different attributes (hard/easy and pos/neg) can be implied by the distribution of gradient norm. The density of examples w.r.t. gradient norm, which we call as \textit{gradient density} for convenient, varies largely as showed in the left of Fig.\ref{fig:method}. The examples with very small gradient norm have a quite large density which is corresponding to the large amount of easy negative examples. Although one easy example has less contribution on the global gradient than a hard example, the total contribution of the huge amount of easy examples can overwhelm the contribution of the minority of hard examples and the training process will be inefficient. Besides, we also discover that the density of examples with very large gradient norm (very hard examples) is slightly larger than the density of the medium examples. And we consider these very hard examples mostly as outliers since they exist stably even when the model is converged. The outliers may affect the stability of model since their gradients may have a large discrepancy from the other common examples.

Inspired by the analysis of gradient norm distribution, we propose a gradient harmonizing mechanism (GHM) to train the one-stage object detection model in an efficient, which focuses on the harmony of gradient contribution of different examples. The GHM first performs statistics on the number of examples with similar attributes w.r.t their gradient density and then attach a harmonizing parameter to the gradient of each example according to the density. The effect of GHM compared with CE and FL is illustrated in the right of Fig.\ref{fig:method}. Training with GHM, the huge amount of cumulated gradient produced by easy examples can be largely down-weighted and the outliers can be relatively down-weighted as well. In the end, the contribution of each kind of examples will be balanced and the training can be more efficient and stable.  

In practice, the modification of gradient can be equivalently implemented by reformulating the loss function, we embed the GHM into the classification loss, which is denoted as GHM-C loss. This loss function is elegantly formulated without many hyper-parameters to tune. Since the gradient density is a statistical variable depending on the examples distribution in a mini-batch, GHM-C is a dynamic loss that can adapt to the change of data distribution in each batch as well as to the updating of model. To showcase the generality of GHM, we also adopt it in the box regression branch as the form of GHM-R loss.

Experiments on the bounding box detection track of the challenging COCO benchmark show that the GHM-C loss has a large gain compared to the traditional cross-entropy loss and slightly surpasses the state-of-the-art Focal Loss. And the GHM-R loss also has better performance than the commonly used smooth $L_1$ loss. The combination of GHM-C and GHM-R attains a new state-of-the-art performance on COCO \textit{tes-dev} set.

Our main contributions are as follows:
\begin{enumerate}
\item We reveal the essential principle behind the significant example imbalance in one-stage detector in term of gradient norm distribution, and propose a novel gradient harmonizing mechanism (GHM) to handle it.
\item We embed the GHM into the loss for classification and regression as GHM-C and GHM-R respectively, which rectify the gradient contribution of examples with different attributes and is robust to hyper-parameters. 
\item Collaborating with GHM, we can easily train a single stage detector without any data sampling strategy and achieve the state-of-the-art result on COCO benchmark.
\end{enumerate}

\section{Related Work}

\subsubsection{Object Detection:}
Object detection is one of the most basic and important task in the field of computer vision. Deep convolutional neural network (CNN) based methods, e.g. \cite{faster,ssd,yolov2,mask}, have become more and more developed and achieved great success in recent years, owing to the significant progress of network architecture such as \cite{vgg,googlenet,resnet,densely}. Advanced object detection frameworks can be divided into two categories: one-stage detector and two-stage detector.

 Most state of the art methods use two-stage detectors, e.g. \cite{fast,faster,li2017zoom,mask,fpn,zeng2018crafting}. They are mainly based on the Region CNN (R-CNN) architecture. These approaches first obtain a manageable number of region proposals called region of interest (RoI) from the nearly infinite candidate regions and then use the network to evaluate each RoI. 

 One-stage detectors have the advantage of simple structures and high speed. SSD \cite{ssd,dssd}, YOLO \cite{yolo,yolov2,yolov3} for generic object detection and RSA~\cite{songbeyond,liu2017recurrent} for face detection have achieved good speed/accuracy trade-off. However, they can hardly surpass the accuracy of two-stage detectors. RetinaNet \cite{focal} is the state of the art one-stage object detector that achieve comparable performance to two-stage detectors. It adopts an architecture modified from RPN \cite{faster} and focuses on addressing the class imbalance during training.

\subsubsection{Object Functions for Object Detector:}
Most detection models use cross entropy based loss function for classification \cite{fast,faster,ssd,rfcn,fpn,mask}. While one-stage detectors face a problem of extreme class imbalance that two-stage detectors do not have. Earlier methods try to use hard example mining methods, e.g. \cite{ohem,casdpm}, but they discard most examples and cannot handle the problem well. Recently the work \cite{focal} reformulate the cross-entropy loss so that easy negatives are down-weighted and the hard examples are unaffected or even up-weighted. 

For stable training of box regression, Fast R-CNN \cite{fast} introduces the smooth $L_1$ loss. This loss reduces the impact of outliers so that the training of model can be more stable. Almost all the following works take the smooth $L_1$ loss as a default for box regression \cite{faster,ssd,rfcn,fpn,mask}. 

The work \cite{dist} tries to improve regression performance by changing the target to a distribution and using a histogram loss to calculate the K-L divergence of prediction and target. The work \cite{grad} balances multi-task losses by dynamically tuning gradient magnitude of different task branches.

Our GHM based loss harmonizes the contribution of examples on the basis of the distribution of their gradient, so that it can handle both the class imbalance and the outliers problem well. It can also adapt the weights to the changing of data distribution in each mini-batch.

\section{Gradient Harmonizing Mechanism}
\label{sec:ghm}

\subsection{Problem Description}
Similar to \cite{focal}, our efforts here are focused on classification in one-stage object detection where the classes (foreground/background) of examples are quite imbalanced. For a candidate box, let $p \in [0,1]$ be the probability predicted by the model and $p^* \in \{0,1\}$ be its ground-truth label for a certain class. Consider the binary cross entropy loss: 
\begin{equation}
\label{eq:ce}
  L_{CE}(p,p^*) = \left\{
    \begin{aligned}
    & -\log(p)  & \text{if } p^* = 1 \\
    & -\log(1-p) & \text{if } p^* = 0 
    \end{aligned}
  \right.
\end{equation}
Let x be the direct output of the model such that $p = \text{sigmoid}(x)$, we have the gradient with regard to x:
\begin{equation}
\begin{aligned}
\frac{\partial L_{CE}}{\partial x} &= \left\{
    \begin{aligned}
    & p - 1  & \text{if } p^* = 1 \\
    & p & \text{if } p^* = 0 
    \end{aligned} 
    \right. \\
    &= p - p^*
\end{aligned}
\end{equation}
We define $g$ as follows:
\begin{equation}
    g = |p - p^*| = \left\{
    \begin{aligned}
    & 1 - p  & \text{if } p^* = 1 \\
    & p & \text{if } p^* = 0 
    \end{aligned}
    \right.
\end{equation}
$g$ equals to the norm of gradient w.r.t $x$. The value of $g$ represents attribute (e.g. easy or hard) of an example and implies the example's impact on the global gradient. Although the strict definition of gradient is on the whole parameter space, which means $g$ is a relative norm of an example's gradient, we call $g$ as gradient norm in this paper for convenience.

\begin{figure}[ht]
\centering
\includegraphics[width=0.8\linewidth]{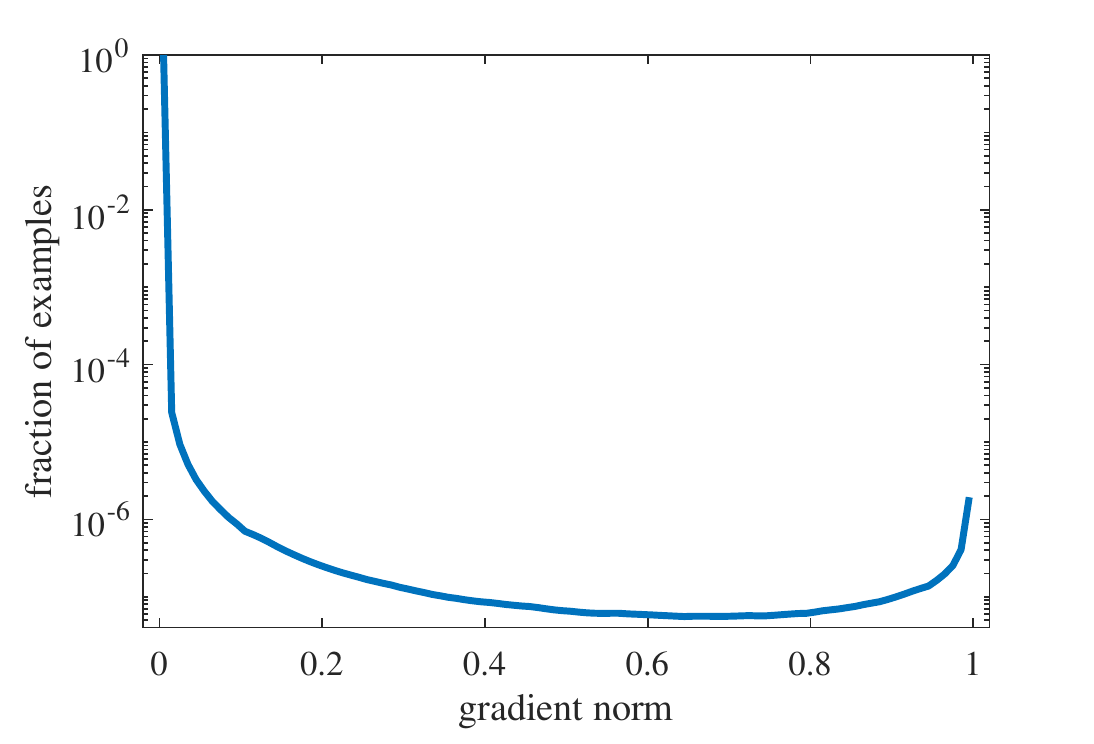}
\caption{The distribution of the gradient norm $g$ from a converged one-stage detection model. Note that the y-axis uses log scale since the number of examples with different gradient norm can differ by orders of magnitude.}
\label{fig:cls_dist_log}
\end{figure}

Fig.\ref{fig:cls_dist_log} shows the distribution of $g$ from a converged one-stage detection model. Since the easy negatives have a dominant number, we use log axis to display the fraction of examples to demonstrate the details of the variance of examples with different attributes. It can be seen that the number of very easy examples is extremely large, which have a great impact on the global gradient. Moreover, we can see that a converged model still can't handle some very hard examples whose number is even larger than the examples with medium difficulty. These very hard examples can be regarded as outliers since their gradient directions tends to vary largely from the gradient directions of the large amount of other examples. That is, if the converged model is forced to learn to classify these outliers better, the classification of the large number of other examples tends to be less accurate.

\subsection{Gradient Density}

To handle the problem of the disharmony of gradient norm distribution, we introduce a harmonizing approach with regard to gradient density. Gradient density function of training examples is formulated as Equation.\ref{eq:gd}:
\begin{equation}
\label{eq:gd}
  GD(g) = \frac{1}{l_\epsilon(g)}\sum_{k=1}^N\delta_\epsilon(g_k, g)
\end{equation}
where $g_k$ is the gradient norm of the k-th example. And
\begin{equation}
  \delta_\epsilon(x, y) = \left\{
    \begin{aligned}
    & 1  & \text{if } y-\frac{\epsilon}{2} <= x < y+\frac{\epsilon}{2} \\
    & 0  & \text{otherwise}
    \end{aligned}
    \right.
\end{equation}

\begin{equation}
l_\epsilon(g) = \min(g+\frac{\epsilon}{2}, 1) - \max(g-\frac{\epsilon}{2}, 0)
\end{equation}
The gradient density of $g$ denotes the number of examples lying in the region centered at $g$ with a length of $\epsilon$ and normalized by the valid length of the region. 

Now we define the gradient density harmonizing parameter as:
\begin{equation}
\label{eq:beta}
  \beta_i = \frac{N}{GD(g_i)}
\end{equation}
where $N$ is the total number of examples. To better comprehend the gradient density harmonizing parameter, we can rewrite it as $\beta_i = \frac{1}{GD(g_i)/N}$. The denominator $GD(g_i)/N$ is a normalizer indicating the fraction of examples with neighborhood gradients to the i-th example. If the examples are uniformly distributed with regard to gradient, $GD(g_i) = N$ for any $g_i$ and each example will have the same $\beta_i = 1$, which means nothing is changed. Otherwise, the examples with large density will be relatively down-weighted by the normalizer.

\subsection{GHM-C Loss}

We embed the GHM into classification loss by regarding $\beta_i$ as the loss weight of the i-th example and the gradient density harmonized form of loss function is:
\begin{equation}
\label{eq:lghm}
    \begin{aligned}
    L_{GHM-C} &= \frac{1}{N}\sum_{i=1}^N\beta_i L_{CE}(p_i, p_i^*) \\
    &= \sum_{i=1}^N\frac{L_{CE}(p_i, p_i^*)}{GD(g_i)} 
    \end{aligned}
\end{equation}

\begin{figure}[ht]
\centering
\includegraphics[width=0.8\linewidth]{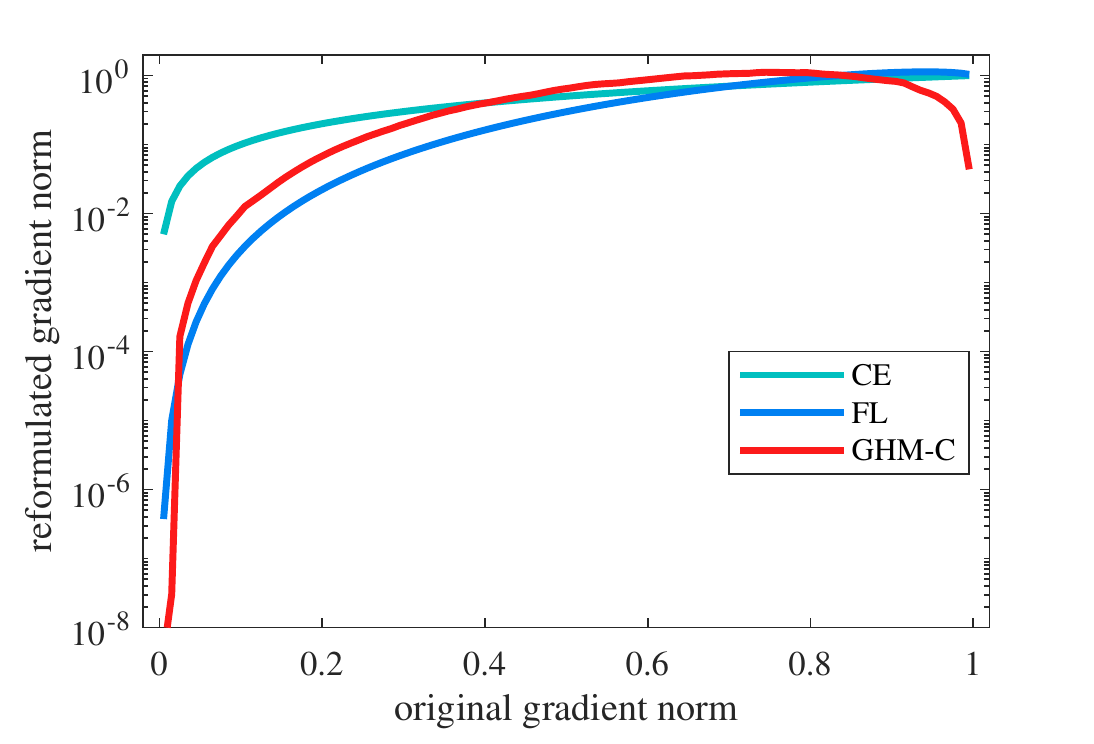}
\caption{Reformulated gradient norm of different loss functions w.r.t the original gradient norm $g$. The y-axis uses log scale to better display the details of FL and GHM-C.}
\label{fig:cls_grads}
\end{figure}

Fig.\ref{fig:cls_grads} illustrates the reformulated gradient norm of different losses. Here we take the original gradient norm of CE, i.e. $g = |p-p^*|$, as the x-axis for convenient view since the density is calculated according to $g$. We can see that the curves of Focal Loss and GHM-C loss have similar trend, which implies that Focal Loss with the best hyper-parameters is similar with uniform gradient harmonizing. Furthermore, GHM-C has one more merit that Focal loss ignores: down-weighting the gradient contribution of outliers.

With our GHM-C loss, the huge number of very easy examples are largely down-weighted and the outliers are slightly down-weighted as well, which simultaneously addresses the attribute imbalance problem and the outliers problem. From the right figure in Fig.\ref{fig:method} we can better see that GHM-C harmonizes the total gradient contribution of different group of examples. Since the gradient density is calculated every iteration, the weights of examples are not fixed w.r.t. $g$ (or $x$) like focal loss but adaptive to current state of model and mini-batch of data. The dynamic property of  GHM-C loss makes the training more efficient and robust.

\subsection{Unit Region Approximation}
\subsubsection{Complexity Analysis:} The naive algorithm to calculate the gradient density values of all examples has a time complexity of $O(N^2)$, which can be easily attained from Equations \ref{eq:gd} and \ref{eq:lghm}. Even parallel computed, each computing unit still bears a computation of $N$. And as far as we know, the best algorithm first sort the examples by gradient norm with a complexity of $O(N\log N)$ and then use a queue to scan the examples and get their density with $O(N)$. This sorting based method can not gain much from parallel computing. Since $N$ of an image in one-stage detector can be $10^5$ or even $10^6$, to directly calculate the gradient density is quite time consuming. So we introduce an alternative approach to approximately attain the gradient density of examples.

\subsubsection{Unit Region:} We divide the range space of $g$ into individual unit regions with a length of $\epsilon$, and there are $M = \frac{1}{\epsilon}$ unit regions. Let $r_j$ be the unit region with index j so that $r_j = [(j-1)\epsilon, j\epsilon)$. Let $R_j$ denote the number of examples lying in $r_j$. We define $ind(g) = t$ s.t. $(t-1)\epsilon <= g < t\epsilon$, which is the index function to the unit region that $g$ lies in. Then we define the approximate gradient density function as:
\begin{equation}
\label{eq:agd}
  \hat{GD}(g) = \frac{R_{ind(g)}}{\epsilon} = R_{ind(g)}M
\end{equation}
Then we have the approximate gradient density harmonizing parameter:
\begin{equation}
\label{eq:abeta}
  \hat{\beta}_i = \frac{N}{\hat{GD}(g_i)}
\end{equation}
Consider the special case where $\epsilon = 1$: there are just one unit region and all examples lie in it, so obviously every $\beta_i = 1$ and each example keep their original gradient contribution.
Finally we have the reformulated loss function:
\begin{equation}
\label{eq:alghm}
  \begin{aligned}
    \hat{L}_{GHM-C} &= \frac{1}{N}\sum_{i=1}^N\hat{\beta}_i L_{CE}(p_i, p_i^*) \\
    &= \sum_{i=1}^N\frac{L_{CE}(p_i, p_i^*)}{\hat{GD}(g_i)} 
    \end{aligned} 
\end{equation}

From Equation. \ref{eq:agd} we can see that the examples lying in the same unit region share the same gradient density. So we can use the algorithm of histogram statistics and the computation of all the gradient density values has a time complexity of $O(MN)$. And parallel computing can be applied so that each computing unit has a computation of $M$. In practice, we can attain good performance with quite small number of unit regions. That is $M$ is fairly small and the calculation of loss is efficient.

\subsubsection{EMA:} Mini-batch statistics based methods usually face a problem: when many extreme data are just sampled in one mini-batch, the statistical result will be a serious noise and the training will be unstable. Exponential moving average (EMA) is a common used method to address this problem, e.g., SGD with momentum \cite{sgd} and Batch Normalization \cite{bn}. Since in the approximation algorithm the gradient densities come from the numbers of examples in the unit regions, we can apply EMA on each unit region to obtain more stable gradient densities for examples. Let $R_j^{(t)}$ be the number of examples in the j-th unit region in the t-th iteration and $S_j^{(t)}$ be the moving averaged number. We have:
\begin{equation}
   S_j^{(t)} = \alpha S_j^{(t-1)} + (1-\alpha) R_j^{(t)}
\end{equation}
where $\alpha$ is the momentum parameter. We use the averaged number $S_j$ to calculate the gradient density instead of $R_j$:
\begin{equation}
  \hat{GD}(g) = \frac{S_{ind(g)}}{\epsilon} = S_{ind(g)}M
\end{equation}
With EMA, the gradient density will be more smooth and insensitive to extreme data.

\subsection{GHM-R Loss}
Consider the parameterized offsets, $t = (t_x, t_y, t_w, t_h)$, predicted by box regression branch and the target offsets, $t^* = (t_x^*, t_y^*, t_w^*, t_h^*)$, computed from ground-truth. The regression loss usually adopts the smooth $L_1$ loss function:
\begin{equation}
\label{eq:reg1}
    L_{reg} = \sum_{i \in \{x,y,w,h\}}SL_1(t_i - t_i^*)
\end{equation}
where
\begin{equation}
\label{eq:smoothl1}
    SL_1(d) = \left\{ 
    \begin{aligned}
        & \frac{d^2}{2\delta} & \text{if }  |d| <= \delta \\
        & |d| - \frac{\delta}{2} & \text{otherwise}
    \end{aligned}
    \right.
\end{equation}
where $\delta$ is the division point between the quadric part and the linear part, and usually set to $1/9$ in practice. 

Since $d = t_i - t_i^*$, the gradient of smooth $L_1$ loss w.r.t $t_i$ can be expressed as:
\begin{equation}
\label{eq:g_smoothl1}
    \frac{\partial SL_1}{\partial t_i} = \frac{\partial SL_1}{\partial d} = \left\{ 
    \begin{aligned}
        & \frac{d}{\delta} & \text{if }  |d| <= \delta \\
        & sgn(d) & \text{otherwise}
    \end{aligned}
    \right.
\end{equation}
where $sgn$ is the sign function. 

Note that all the examples with $|d|$ larger than the division point have the same gradient norm $|\frac{\partial SL_1}{\partial t_i}| = 1$, which makes the distinguishing of examples with different attributes impossible if depending on the gradient norm. An alternative choice is directly using $|d|$ as the measurement of different attributes, but the new problem is $|d|$ can reach to infinite in theory and the unit region approximation can not be implemented.

To conveniently apply  GHM on regression loss, we first modify the traditional $SL_1$ loss into a more elegant form:
\begin{equation}
\label{eq:asl1}
    ASL_1(d) = \sqrt{d^2 + \mu^2} - \mu
\end{equation}
This loss shares similar property with $SL_1$ loss: when d is small it approximates a quadric function ($L_2$ loss) and when d is large is approximate a linear function ($L_1$ loss). We denote the modified loss function as Authentic Smooth $L_1$ ($ASL_1$) loss for its good property of authentic smoothness, which means all the degrees of derivatives are existed and continuous. In contrast, the second derivative of smooth $L_1$ loss does not exist at the point $d = \delta$. Furthermore, the $ASL_1$ loss has an elegant form of gradient w.r.t $d$:
\begin{equation}
\label{eq:g_asl1}
    \frac{\partial ASL_1}{\partial d} = \frac{d}{\sqrt{d^2 + \mu^2}}
\end{equation}
The range of the gradient is just $[0,1)$, so the calculation of density in unit regions for $ASL_1$ loss in regression is as convenient as CE loss in classification. In practice, we set $\mu = 0.02$ for $ASL_1$ loss to keep the same performance with $SL_1$ loss.

\begin{figure}[ht]
\centering
\includegraphics[width=0.8\linewidth]{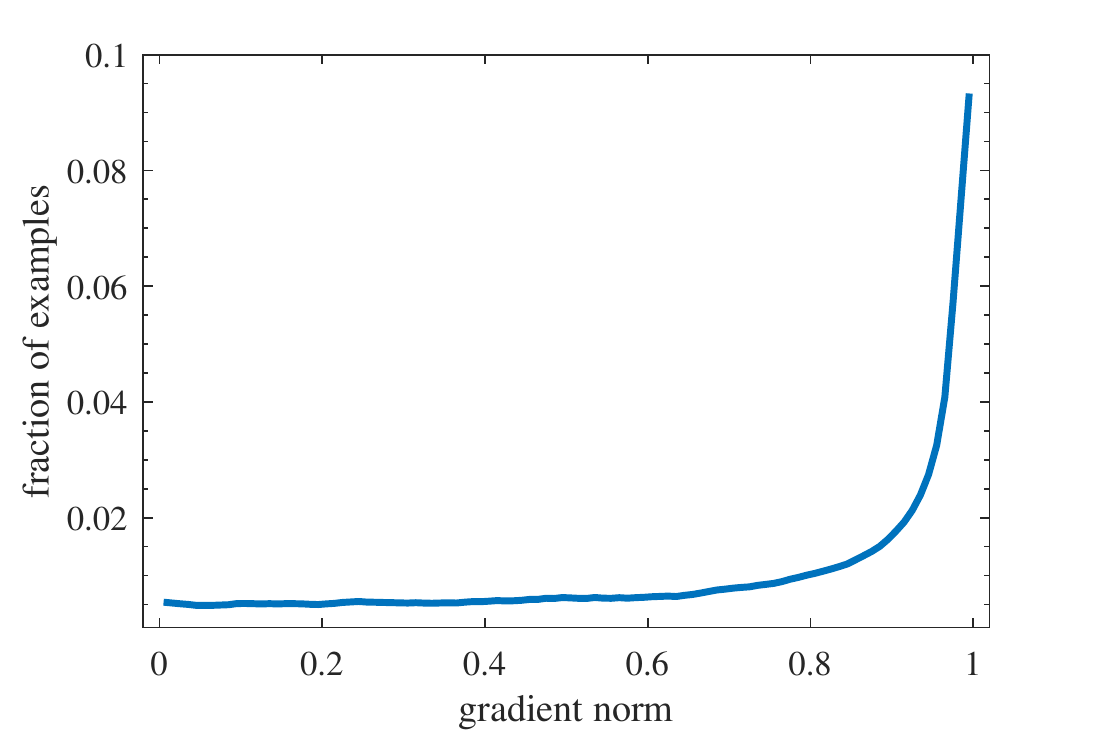}
\caption{The distribution of the gradient norm $gr$ for $ASL_1$ loss.}
\label{fig:reg_dist}
\end{figure}

We define $gr = |\frac{d}{\sqrt{d^2 + \mu^2}}|$ as the gradient norm of $ASL_1$ loss and the gradient distribution of a converged model is illustrated in Fig.\ref{fig:reg_dist}
We can see that there are large number of outliers. Note that the regression is only performed on the positive examples so it is reasonable for the different distribution trend between classification and regression. Above all, we can apply  GHM on regression loss:
\begin{equation}
\label{eq:ghm-r}
    \begin{aligned}
    L_{GHM-R} &= \frac{1}{N}\sum_{i=1}^N\beta_i ASL_1(d_i) \\
    &= \sum_{i=1}^N\frac{ASL_1(d_i)}{GD(gr_i)}
    \end{aligned}
\end{equation}

\begin{figure}[ht]
\centering
\includegraphics[width=0.8\linewidth]{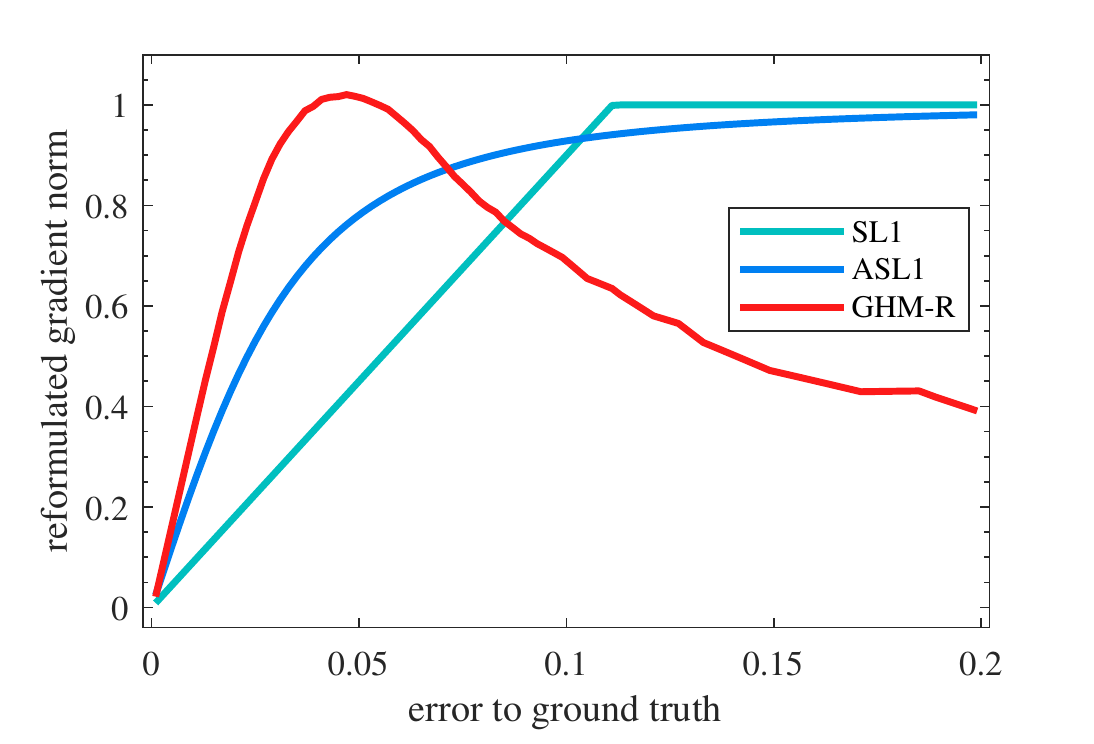}
\caption{Comparison of the reformulated gradient contributions of different regression losses w.r.t the value of $|d|$, i.e. the error to ground-truth.}
\label{fig:reg_grads}
\end{figure}

The reformulated gradient contribution of $SL_1$ loss, $ASL_1$ loss and GHM-R loss in Fig.\ref{fig:reg_grads}. The x-axis adopts $|d|$ for convenient comparison.

We emphasize that in box regression not all the ``easy examples'' are unimportant. An easy example in classification is usually a background region with a very low predicted probability and will be definitely excluded from the final candidates. Thus the improvement of this kind of examples makes nearly no contribution to the precision. But in box regression, an easy example still has deviation from the ground truth location. Better prediction of any example will directly improve the quality of the final candidates. 
Moreover, advanced datasets care more about the localization accuracy. For example, COCO \cite{coco} takes the average AP from the IoU threshold 0.5 to 0.95 as the metric to evaluate an algorithm. In this metric, the some of the so called easy examples (those having small errors) are also important because reducing the errors of them can directly improve the AP at high threshold (e.g. AP@IoU=0.75). 

Our GHM-R loss can harmonize the contribution of easy and hard examples for box regression by up-weighting the important part of easy examples and down-weighting the outliers.
Experiments show that it can attain better performance than $SL_1$ and $ASL_1$.

\section{Experiments}
We evaluate our approach on the challenging COCO benchmark \cite{coco}. For training, we follow the common used practice \cite{mask,focal} to divide the 40k validation set into a 35k subset and a 5k subset. The union of the 35k validation subset and the whole 80k training set are used for training together and denoted as \textit{trainval35k} set. The 5k validation subset is denoted as \textit{minival} set and our ablation study is performed on it. While our main results are reported on the \textit{test-dev} set.

\subsection{Implementation Details}
\subsubsection{Network Setting:} We use RetinaNet \cite{focal} as  network architecture and all the experiments adopt ResNet \cite{resnet} as backbone with Feature Pyramid Network (FPN) \cite{fpn} structure. Anchors use 3 scales and 3 aspect ratios for convenient comparison with focal loss. The input image scale is set as 800 pixel for all experiments. For all ablation studies, ResNet-50 is used. While the final model evaluated on \textit{test-dev} adopts ResNeXt-101 \cite{resnext}. In contrast to focal loss, our approach doesn't need a specialized bias initialization.

\subsubsection{Optimization:} All the models are optimized by the common used SGD algorithm. We train the models on 8 GPUs with 2 images on each GPU so that the effective mini-batch size is 16. All models are trained for 14 epochs with an initial learning rate of 0.01, which is decreased by a factor 0.1 at the 9th epoch and again at the 12th epoch. We also use a weight decay parameter of 0.0001 and a momentum parameter of 0.9. The only data augmentation operation is horizontal image flipping. For the EMA used in gradient density calculation, we use $\alpha=0.75$ for all experiments since the results are insensitive to the exact value of $\alpha$.

\subsection{GHM-C Loss}
To focus on the effect of  GHM-C loss function, experiments in this section all adopt smooth $L_1$ loss function with $\delta=1/9$ for the box regression branch. 

\subsubsection{Baseline:} We have trained a model with the standard cross entropy loss as the baseline. The standard initialization will lead to quick divergence, so we follow focal loss \cite{focal} to initialize the bias term of the last layer to $b = -\log((1-\pi)/\pi)$ with $\pi=0.01$ to avoid divergence. However with the specialized initialization the loss of classification is very small, so we up-weight the classification loss by 20 to make the begging loss value reasonable (the begging classification loss value is around 1 now). But when the model converge, the classification loss is still very small and we finally obtain a model with an Average Precision (AP) of 28.6.

\subsubsection{Number of Unit Region}
Table.\ref{tab:num} shows the results of varying $M$ which is the number of unit regions. EMA is not applied here. When $M$ is too small, the density can not have a good variation over different gradient norm and the performance is not so good. So we can gain more when $M$ increases when $M$ is not large. However $M$ is not necessarily very large, when $M = 30$, the GHM-C loss yields a large enough improvement over  baseline.
\begin{table}[!ht]
\begin{center}
\begin{tabular}{| c | c  c  c  c  c  c |}
\hline
M & AP & $\text{AP}_{.5}$ & $\text{AP}_{.75}$ & $\text{AP}_{S}$ & $\text{AP}_{M}$ & $\text{AP}_{L}$ \\
\hline
5 & 33.4 & 51.7 & 35.6 & 18.6 & 36.8 & 45.7 \\
10 & 34.6 & 53.9 & 36.5 & 19.5 & 37.1 & 46.1 \\
20 & 35.2 & 54.4 & 36.9 & 19.4 & 38.4 & 46.3 \\
30 & \textbf{35.8} & \textbf{55.5} & \textbf{38.1} & \textbf{19.6} & \textbf{39.6} & \textbf{46.7} \\
40 & 35.4 & 54.8 & 36.3 & 19.5 & 38.5 & 46.3 \\
\hline 
\end{tabular}
\caption{Results of varying number of unit regions for GHM-C loss.}
\label{tab:num}
\end{center}
\end{table}

\subsubsection{Speed:} Since our approach is a loss function, it doesn't change the time for inference. For training, a small $M$ of 30 is enough to attain good performance, so time consumed by gradient density calculation is not long. Table.\ref{tab:speed} shows the average time for each iteration during training as well as average precision. Here ``GHM-C Standard'' is implemented using the original definition of gradient density and ``GHM-C RU'' represents the implementation of region unit approximation algorithm. The experiments are performed on 1080Ti GPUs. We can see that our region unit approximation algorithm speed up the training by magnitudes with negligible harm to performance. While compared with CE, the slow down of  GHM-C loss is also acceptable. Since our loss is not fully GPU implemented now, there is still room for improvement.
\begin{table}[!ht]
\begin{center}
\begin{tabular}{| c | c | c |}
\hline
method & AP & average time per iteration (s) \\
\hline
standard CE & 28.6 & 0.566 \\
GHM-C Standard & 35.9 & 13.675 \\
GHM-C RU & 35.8 & 0.824 \\
\hline 
\end{tabular}
\caption{The comparison of training speed as well as AP.}
\label{tab:speed}
\end{center}
\end{table}

\begin{table*}[t]
\begin{tabular}{| c | c | c c c | c c c|}
\hline
 method & network
 & AP & AP$_{50}$ & AP$_{75}$
 & AP$_S$ & AP$_M$ &  AP$_L$\\ [.1em]
\hline
 ~Faster RCNN \cite{faster} & FPN-ResNet-101
  & 36.2 & 59.1 & 39.0 & 18.2 & 39.0 & 48.2\\
 ~Mask RCNN \cite{mask} & FPN-ResNet-101
  & 38.2 & 60.3 & 41.7 & 20.1 & 41.1 & 50.2\\
 ~Mask RCNN \cite{mask} & FPN-ResNeXt-101
  & 39.8 & 62.3 & 43.4 & 22.1 & 43.2 & 51.2\\
\hline
 ~YOLOv3 \cite{yolov3} & DarkNet-53 
  & 33.0 & 57.9 & 34.4 & 18.3 & 35.4 & 41.9 \\
 ~DSSD513 \cite{dssd} & DSSD-ResNet-101
  & 33.2 & 53.3 & 35.2 & 13.0 & 35.4 & 51.1 \\
 ~Focal Loss \cite{focal} & RetinaNet-FPN-ResNet-101
  & 39.1 & 59.1 & 42.3 & 21.8 & 42.7 & 50.2 \\
 ~Focal Loss \cite{focal} & RetinaNet-FPN-ResNeXt-101
  & 40.8 & 61.1 & 44.1 & \textbf{24.1} & 44.2 & 51.2 \\
\hline
 ~GHM-C + GHM-R (ours) & RetinaNet-FPN-ResNet-101
  & 39.9 & 60.8 & 42.5 & 20.3 & 43.6 & 54.1 \\
 ~GHM-C + GHM-R (ours) & RetinaNet-FPN-ResNeXt-101
  & \textbf{41.6} & \textbf{62.8} & \textbf{44.2} & 22.3 & \textbf{45.1} & \textbf{55.3} \\
\hline
\end{tabular}
\caption{Comparison with state-of-the-art methods (single model) on COCO \textit{test-dev} set.}
\label{tab:main}
\end{table*}

\subsubsection{Comparison with Other Methods:} Table.\ref{tab:comp} shows the results using our loss compared with other loss functions or sampling strategy. Since the reported results on $minival$ of models using focal loss is trained with the input image scale of 600 pixels, for fair comparison we have re-trained a focal loss using a scale of 800 pixels and keep the best parameters of focal loss.
We can see our loss has slightly better performance than focal loss.
\begin{table}[!ht]
\begin{center}
\begin{tabular}{| c | c  c  c  c  c  c |}
\hline
method & AP & $\text{AP}_{.5}$ & $\text{AP}_{.75}$ & $\text{AP}_{S}$ & $\text{AP}_{M}$ & $\text{AP}_{L}$ \\
\hline
CE & 28.6 & 43.3 & 30.7 & 11.4 & 30.7 & 40.7 \\
OHEM & 31.1 & 47.2 & 33.2 & - & - & - \\
FL & 35.6 & \textbf{55.6} & \textbf{38.2} & 19.1 & 39.2 & 46.3 \\
\hline
GHM-C & \textbf{35.8} & 55.5 & 38.1 & \textbf{19.6} & \textbf{39.6} & \textbf{46.7} \\
\hline
\end{tabular}
\caption{Comparison of other loss functions. Note that the 'OHEM' is trained with ResNet-101 while others are trained with ResNet-50.}
\label{tab:comp}
\end{center}
\end{table}

\subsection{GHM-R Loss}
\subsubsection{Comparison with Other Losses:} The experiments here adopt the best configuration of GHM-C loss for the classification branch. So the first baseline is the model (trained using $SL_1$ loss) with an AP of 35.8 showed in GHM-C loss experiments. We adopts $\mu=0.02$ for $ASL_1$ loss to get comparable results with $SL_1$ loss and obtain a fair baseline for GHM-R loss. Table.\ref{tab:reg1} shows the results of the baseline $SL_1$ and $ASL_1$ loss as well as  GHM-R loss. We can see a gain of 0.7 mAP based on the $ASL_1$ loss. Table.\ref{tab:regthr} shows the details of AP at different IoU thresholds.  GHM-R loss slightly lowers the AP@IoU=0.5 but gains when the threshold is higher, which demonstrates our proposition that the so called easy examples in regression is important for accurate localization.
\begin{table}[!ht]
\begin{center}
\begin{tabular}{| c | c  c  c  c  c  c |}
\hline
method & AP & $\text{AP}_{.5}$ & $\text{AP}_{.75}$ & $\text{AP}_{S}$ & $\text{AP}_{M}$ & $\text{AP}_{L}$ \\
\hline
$SL_1$ & 35.8 & \textbf{55.5} & 38.1 & 19.6 & 39.6 & 46.7 \\
$ASL_1$ & 35.7 & 55.0 & 38.1 & 19.7 & 39.7 & 45.9 \\
GHM-R & \textbf{36.4} & 54.6 & \textbf{38.7} & \textbf{20.5} & \textbf{40.6} & \textbf{47.8} \\
\hline 
\end{tabular}
\caption{Comparison of different loss functions for regression.}
\label{tab:reg1}
\end{center}
\end{table}

\begin{table}[!ht]
\begin{center}
\begin{tabular}{| c | c  c  c  c  c  c |}
\hline
method & AP & $\text{AP}_{.5}$ & $\text{AP}_{.6}$ & $\text{AP}_{.7}$ & $\text{AP}_{.8}$ & $\text{AP}_{.9}$ \\
\hline
$SL_1$ & 35.8 & \textbf{55.5} & 51.2 & 43.4 & 31.4 & 11.9 \\
$ASL_1$ & 35.7 & 55.0 & 51.1 & 43.5 & 31.5 & 12.1 \\
GHM-R & \textbf{36.4} & 54.6 & \textbf{51.4} & \textbf{44.0} & \textbf{32.2} & \textbf{13.1} \\
\hline
\end{tabular}
\caption{Comparison of AP at different IoU thresholds.}
\label{tab:regthr}
\end{center}
\end{table}

\subsubsection{Two-Stage Detector:}  GHM-R loss for regression is not limited to one-stage detectors. So we have done experiments to verify the effect on two-stage detectors. Our baseline method is faster-RCNN with Res50-FPN model using $\text{SL}_1$ loss for box regression. Table.\ref{tab:reg2} shows that  GHM-R loss works for two-stage detector as well as one-stage detector.
\begin{table}[!ht]
\begin{center}
\begin{tabular}{| c | c  c  c  c  c  c |}
\hline
method & AP & $\text{AP}_{.5}$ & $\text{AP}_{.75}$ & $\text{AP}_{S}$ & $\text{AP}_{M}$ & $\text{AP}_{L}$ \\
\hline
$\text{SL}_1$ & 36.4 & 58.7 & 38.8 & 21.1 & 39.6 & 47.0 \\
GHM-R & \textbf{37.4} & \textbf{58.9} & \textbf{39.9} & \textbf{21.8} & \textbf{40.8} & \textbf{48.8} \\
\hline
\end{tabular}
\caption{Comparison of regression loss functions on two-stage detector.}
\label{tab:reg2}
\end{center}
\end{table}

\subsection{Main Results}
We use the 32x8d FPN-ResNext101 backbone and RetinaNet model with  GHM-C loss for classification and GHM-R loss for box regression. The experiments are performed on \textit{test-dev} set. Table.\ref{tab:main} shows our main result compared with state-of-the-art methods. Our approach achieves excellent performance and outperforms focal loss in most metrics.

\section{Conclusion and Discussion}
In this work, we focus on the two imbalance problems in single-stage detectors and summarize these two problems to the disharmony in gradient density with regard to the difficulty of samples. Two loss functions, GHM-C and GHM-R are proposed to conquer the disharmony in classification and bounding box regression respectively. Experiments show that the collaborate with GHM, the performance of single-stage detector can easily surpass modern state-of-the-art two-stage detectors like FPN and Mask-RCNN with the same network backbone.

Despite of the improvement of select uniform distribution to be the target, we still hold the opinion that the optimal distribution of gradient is hard to define and requires further research.

\section{Acknowledgments}
We sincerely appreciate the technical and GPU support from Mr. Changbao Wang, Quanquan Li and Junjie Yan at Sensetime Research. And we also acknowledge the early discussion with Prof. Wanli Ouyang from University of Sydney.

\bibliography{mybib}

\begin{thebibliography}{}

\bibitem[\protect\citeauthoryear{Chen \bgroup et al\mbox.\egroup }{2017}]{grad}
Chen, Z.; Badrinarayanan, V.; Lee, C.-Y.; and Rabinovich, A.
\newblock 2017.
\newblock Gradnorm: Gradient normalization for adaptive loss balancing in deep
  multitask networks.
\newblock {\em arXiv preprint arXiv:1711.02257}.

\bibitem[\protect\citeauthoryear{Dai \bgroup et al\mbox.\egroup }{2016}]{rfcn}
Dai, J.; Li, Y.; He, K.; and Sun, J.
\newblock 2016.
\newblock R-fcn: Object detection via region-based fully convolutional
  networks.
\newblock In {\em Advances in neural information processing systems},
  379--387.

\bibitem[\protect\citeauthoryear{Felzenszwalb, Girshick, and
  McAllester}{2010}]{casdpm}
Felzenszwalb, P.~F.; Girshick, R.~B.; and McAllester, D.
\newblock 2010.
\newblock Cascade object detection with deformable part models.
\newblock In {\em Computer vision and pattern recognition (CVPR), 2010 IEEE
  conference on},  2241--2248.
\newblock IEEE.

\bibitem[\protect\citeauthoryear{Fu \bgroup et al\mbox.\egroup }{2017}]{dssd}
Fu, C.-Y.; Liu, W.; Ranga, A.; Tyagi, A.; and Berg, A.~C.
\newblock 2017.
\newblock Dssd: Deconvolutional single shot detector.
\newblock {\em arXiv preprint arXiv:1701.06659}.

\bibitem[\protect\citeauthoryear{Girshick}{2015}]{fast}
Girshick, R.
\newblock 2015.
\newblock Fast r-cnn.
\newblock In {\em Proceedings of the IEEE international conference on computer
  vision},  1440--1448.

\bibitem[\protect\citeauthoryear{He \bgroup et al\mbox.\egroup }{2016}]{resnet}
He, K.; Zhang, X.; Ren, S.; and Sun, J.
\newblock 2016.
\newblock Deep residual learning for image recognition.
\newblock In {\em Proceedings of the IEEE conference on computer vision and
  pattern recognition},  770--778.

\bibitem[\protect\citeauthoryear{He \bgroup et al\mbox.\egroup }{2017}]{mask}
He, K.; Gkioxari, G.; Doll{\'a}r, P.; and Girshick, R.
\newblock 2017.
\newblock Mask r-cnn.
\newblock In {\em Computer Vision (ICCV), 2017 IEEE International Conference
  on},  2980--2988.
\newblock IEEE.

\bibitem[\protect\citeauthoryear{Hu, Shen, and Sun}{2017}]{resnext}
Hu, J.; Shen, L.; and Sun, G.
\newblock 2017.
\newblock Squeeze-and-excitation networks.
\newblock {\em arXiv preprint arXiv:1709.01507} 7.

\bibitem[\protect\citeauthoryear{Huang \bgroup et al\mbox.\egroup
  }{2017}]{densely}
Huang, G.; Liu, Z.; Van Der~Maaten, L.; and Weinberger, K.~Q.
\newblock 2017.
\newblock Densely connected convolutional networks.
\newblock In {\em CVPR}, volume~1, ~3.

\bibitem[\protect\citeauthoryear{Imani and White}{2018}]{dist}
Imani, E., and White, M.
\newblock 2018.
\newblock Improving regression performance with distributional losses.
\newblock {\em arXiv preprint arXiv:1806.04613}.

\bibitem[\protect\citeauthoryear{Ioffe and Szegedy}{2015}]{bn}
Ioffe, S., and Szegedy, C.
\newblock 2015.
\newblock Batch normalization: Accelerating deep network training by reducing
  internal covariate shift.
\newblock {\em arXiv preprint arXiv:1502.03167}.

\bibitem[\protect\citeauthoryear{Li \bgroup et al\mbox.\egroup
  }{2017}]{li2017zoom}
Li, H.; Liu, Y.; Ouyang, W.; and Wang, X.
\newblock 2017.
\newblock Zoom out-and-in network with map attention decision for region
  proposal and object detection.
\newblock {\em International Journal of Computer Vision}  1--14.

\bibitem[\protect\citeauthoryear{Lin \bgroup et al\mbox.\egroup }{2014}]{coco}
Lin, T.-Y.; Maire, M.; Belongie, S.; Hays, J.; Perona, P.; Ramanan, D.;
  Doll{\'a}r, P.; and Zitnick, C.~L.
\newblock 2014.
\newblock Microsoft coco: Common objects in context.
\newblock In {\em European conference on computer vision},  740--755.
\newblock Springer.

\bibitem[\protect\citeauthoryear{Lin \bgroup et al\mbox.\egroup }{2017a}]{fpn}
Lin, T.-Y.; Doll{\'a}r, P.; Girshick, R.~B.; He, K.; Hariharan, B.; and
  Belongie, S.~J.
\newblock 2017a.
\newblock Feature pyramid networks for object detection.
\newblock In {\em CVPR}, volume~1, ~4.

\bibitem[\protect\citeauthoryear{Lin \bgroup et al\mbox.\egroup
  }{2017b}]{focal}
Lin, T.-Y.; Goyal, P.; Girshick, R.; He, K.; and Dollar, P.
\newblock 2017b.
\newblock Focal loss for dense object detection.
\newblock In {\em 2017 IEEE International Conference on Computer Vision
  (ICCV)},  2999--3007.
\newblock IEEE.

\bibitem[\protect\citeauthoryear{Liu \bgroup et al\mbox.\egroup }{2016}]{ssd}
Liu, W.; Anguelov, D.; Erhan, D.; Szegedy, C.; Reed, S.; Fu, C.-Y.; and Berg,
  A.~C.
\newblock 2016.
\newblock Ssd: Single shot multibox detector.
\newblock In {\em European conference on computer vision},  21--37.
\newblock Springer.

\bibitem[\protect\citeauthoryear{Liu \bgroup et al\mbox.\egroup
  }{2017}]{liu2017recurrent}
Liu, Y.; Li, H.; Yan, J.; Wei, F.; Wang, X.; and Tang, X.
\newblock 2017.
\newblock Recurrent scale approximation for object detection in cnn.
\newblock In {\em IEEE international conference on computer vision}, volume~5.

\bibitem[\protect\citeauthoryear{Redmon and Farhadi}{2017}]{yolov2}
Redmon, J., and Farhadi, A.
\newblock 2017.
\newblock Yolo9000: Better, faster, stronger.
\newblock In {\em 2017 IEEE Conference on Computer Vision and Pattern
  Recognition (CVPR)},  6517--6525.
\newblock IEEE.

\bibitem[\protect\citeauthoryear{Redmon and Farhadi}{2018}]{yolov3}
Redmon, J., and Farhadi, A.
\newblock 2018.
\newblock Yolov3: An incremental improvement.
\newblock {\em arXiv preprint arXiv:1804.02767}.

\bibitem[\protect\citeauthoryear{Redmon \bgroup et al\mbox.\egroup
  }{2016}]{yolo}
Redmon, J.; Divvala, S.; Girshick, R.; and Farhadi, A.
\newblock 2016.
\newblock You only look once: Unified, real-time object detection.
\newblock In {\em Proceedings of the IEEE conference on computer vision and
  pattern recognition},  779--788.

\bibitem[\protect\citeauthoryear{Ren \bgroup et al\mbox.\egroup
  }{2015}]{faster}
Ren, S.; He, K.; Girshick, R.; and Sun, J.
\newblock 2015.
\newblock Faster r-cnn: Towards real-time object detection with region proposal
  networks.
\newblock In {\em Advances in neural information processing systems},  91--99.

\bibitem[\protect\citeauthoryear{Shrivastava, Gupta, and Girshick}{2016}]{ohem}
Shrivastava, A.; Gupta, A.; and Girshick, R.
\newblock 2016.
\newblock Training region-based object detectors with online hard example
  mining.
\newblock In {\em Proceedings of the IEEE Conference on Computer Vision and
  Pattern Recognition},  761--769.

\bibitem[\protect\citeauthoryear{Simonyan and Zisserman}{2014}]{vgg}
Simonyan, K., and Zisserman, A.
\newblock 2014.
\newblock Very deep convolutional networks for large-scale image recognition.
\newblock {\em CoRR} abs/1409.1556.

\bibitem[\protect\citeauthoryear{Song \bgroup et al\mbox.\egroup
  }{}]{songbeyond}
Song, G.; Liu, Y.; Jiang, M.; Wang, Y.; Yan, J.; and Leng, B.
\newblock Beyond trade-off: Accelerate fcn-based face detector with higher
  accuracy.
\newblock In {\em 2018 IEEE Conference on Computer Vision and Pattern
  Recognition (CVPR)}.

\bibitem[\protect\citeauthoryear{Sutskever \bgroup et al\mbox.\egroup
  }{2013}]{sgd}
Sutskever, I.; Martens, J.; Dahl, G.; and Hinton, G.
\newblock 2013.
\newblock On the importance of initialization and momentum in deep learning.
\newblock In {\em International conference on machine learning},  1139--1147.

\bibitem[\protect\citeauthoryear{Szegedy \bgroup et al\mbox.\egroup
  }{2016}]{googlenet}
Szegedy, C.; Vanhoucke, V.; Ioffe, S.; Shlens, J.; and Wojna, Z.
\newblock 2016.
\newblock Rethinking the inception architecture for computer vision.
\newblock In {\em Proceedings of the IEEE conference on computer vision and
  pattern recognition},  2818--2826.

\bibitem[\protect\citeauthoryear{Zeng \bgroup et al\mbox.\egroup
  }{2018}]{zeng2018crafting}
Zeng, X.; Ouyang, W.; Yan, J.; Li, H.; Xiao, T.; Wang, K.; Liu, Y.; Zhou, Y.;
  Yang, B.; Wang, Z.; et~al.
\newblock 2018.
\newblock Crafting gbd-net for object detection.
\newblock {\em IEEE transactions on pattern analysis and machine intelligence}
  40(9):2109--2123.

\end{thebibliography}
\bibliographystyle{aaai}

\end{document}